\documentclass[sigconf,nonacm,preprint]{acmart}

\usepackage{silence}
\usepackage{listings}
\usepackage{float}
\usepackage{graphicx}
\usepackage{booktabs}
\usepackage[dvipsnames]{xcolor} 
\usepackage{pgfplots}
\usepackage{pgf-pie}
\usepackage{hyperref}
\usepackage{multirow}

\usepackage{url}
\usepackage{enumitem}
\usepackage{tikz}
\usetikzlibrary{positioning,arrows.meta,calc,fit}

\usepackage{xcolor}
\usepackage{subcaption} 

\usepackage{listings} 

\definecolor{exactmatch}{RGB}{49,76,109}
\definecolor{semanticllm}{RGB}{143,100,118}
\definecolor{fuzzytol}{RGB}{85,118,95}

\definecolor{inputpdf}{RGB}{42,75,115}
\definecolor{outputjson}{RGB}{150,95,118}
\definecolor{comprgreen}{RGB}{60,110,80}

\definecolor{gold}{RGB}{170,130,35} 
\definecolor{purple}{RGB}{106,13,173}      

\pgfplotsset{compat=1.18}
\AtBeginDocument{%
  }
\setcopyright{acmlicensed}
\copyrightyear{2026}
\acmYear{2026}
\acmDOI{XXXXXXX.XXXXXXX}

\acmConference[KDD '26]{Proceedings of the 32nd ACM SIGKDD Conference on Knowledge Discovery and Data Mining}{August 9--13, 2026}{Jeju, Korea}
\acmISBN{978-1-4503-XXXX-X/2026/08}

\title{ExtractBench: A Benchmark and Evaluation Methodology for Complex Structured Extraction}

\newcommand{\contextualauthor}[2]{%
  \author{#1}
  \email{#2}
  \affiliation{%
    \institution{Contextual AI}
    \city{Mountain View}
    \state{CA}
    \country{USA}
  }
}

\author{Nick Ferguson}
\authornote{Equal contribution.}
\email{nick.ferguson@collab.contextual.ai}
\author{Josh Pennington}
\authornotemark[1]
\email{josh@contextual.ai}
\affiliation{%
  \institution{Contextual AI}
  \city{Mountain View}
  \state{CA}
  \country{USA}
}
\contextualauthor{Narek Beghian}{narek.beghian@contextual.ai}
\contextualauthor{Aravind Mohan}{ara@contextual.ai}
\contextualauthor{Douwe Kiela}{douwe@contextual.ai}
\contextualauthor{Sheshansh Agrawal}{sheshansh@contextual.ai}
\author{Thien Hang Nguyen}
\authornote{Corresponding author.}
\email{thien.nguyen@contextual.ai}
\affiliation{%
  \institution{Contextual AI}
  \city{Mountain View}
  \state{CA}
  \country{USA}
}

\renewcommand{\shortauthors}{Ferguson et al.}
\newcommand{\githubrepo}{{\url{https://github.com/ContextualAI/extract-bench}}}

\begin{document}

\begin{abstract}
Unstructured documents like PDFs contain valuable structured information, but downstream systems require this data in reliable, standardized formats. 
LLMs are increasingly deployed to automate this extraction, making accuracy and reliability paramount.
However, progress is bottlenecked by two gaps. First, no end-to-end benchmark evaluates PDF-to-JSON extraction under enterprise-scale schema breadth. Second, no principled methodology captures the semantics of nested extraction, where fields demand different notions of correctness (exact match for identifiers, tolerance for quantities, semantic equivalence for names), arrays require alignment, and omission must be distinguished from hallucination.
We address both gaps with ExtractBench, an open-source benchmark and evaluation framework for PDF-to-JSON structured extraction. The benchmark pairs 35 PDF documents with JSON Schemas and human-annotated gold labels across economically valuable domains, yielding 12,867 evaluatable fields spanning schema complexities from tens to hundreds of fields. The evaluation framework treats the schema as an executable specification: each field declares its scoring metric.
Baseline evaluations reveal that frontier models (GPT-5/5.2, Gemini-3 Flash/Pro, Claude 4.5 Opus/Sonnet) remain unreliable on realistic schemas. Performance degrades sharply with schema breadth, culminating in \textit{0\% valid output} on a 369-field financial reporting schema across all tested models. We release ExtractBench at \githubrepo{}.
\end{abstract}



\maketitle


\section{Introduction}
\label{sec:intro}
\begin{figure}[t]
\centering
\includegraphics[width=\columnwidth]{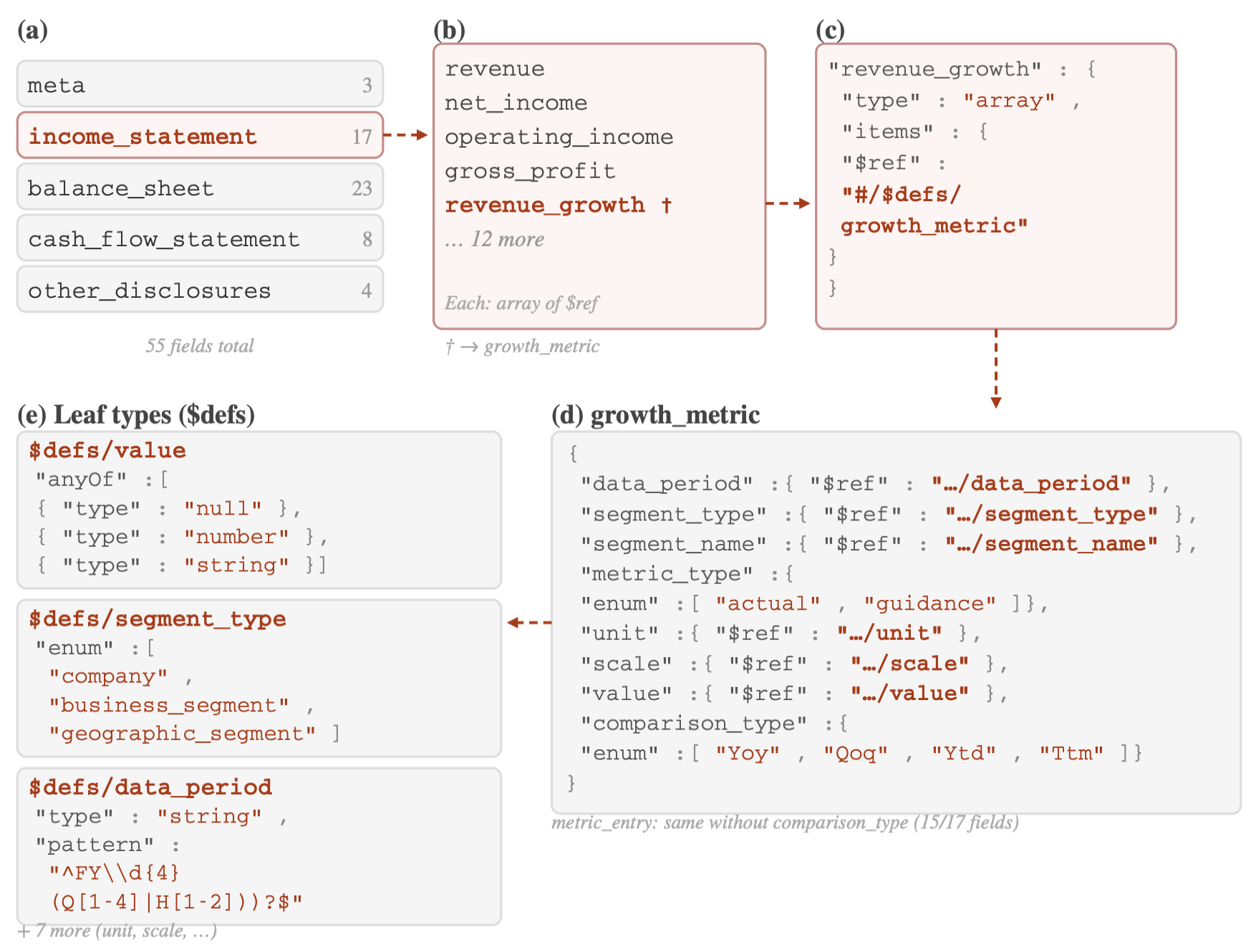}
\caption{Overview of the financial data schema structure showing (a) main sections, (b) key metric fields, (c) JSON schema reference example, (d) growth metric definition with 17 fields, and (e) leaf type definitions.}
\label{fig:schema}
\end{figure}

Enterprises run on structured data, but much of what they need starts as unstructured documents -- contracts, filings, reports. As LLMs move into production automation, structured extraction becomes a core requirement, not a niche task.

At first glance, extracting JSON from a document appears straightforward: provide the PDF and a JSON Schema and ask a model to populate it. However, in practice, real enterprise schemas are not simple (see for example Figure \ref{fig:schema}). They mix many field types (identifiers, numbers, free text, optional fields), are deeply nested, and array types create unpredictable complexity. There are two immediate challenges. First, long documents and large schemas force models to produce long, precisely structured outputs, which are brittle to truncation and formatting errors. Second, syntactically valid output can be wrong in subtle ways. Correctness is field-dependent: IDs need exact match, quantities need tolerance, names need semantic comparison, and arrays demand careful handling of ordering and per-item matching. 

Existing evaluations benchmark and methodology isolate only parts of the problem. There are 2 main gaps today:
\begin{itemize}[topsep=0pt]
\item \emph{Gap~1 (Benchmark):} Document understanding benchmarks focus on span- or entity-level extraction with relatively small schema. Structured output benchmarks test schema conformance under clean text prompts but not extraction correctness. Nested JSON extraction benchmarks use text inputs with modest schema breadth. No existing benchmark jointly covers extraction from PDFs, enterprise-scale JSON schemas, and fine-grained key-level evaluation.
\item \emph{Gap~2 (Methodology):} Existing frameworks apply a single recursive comparison that treats every field identically -- exact equality for primitives, position-dependent matching for arrays of objects, and set overlap for arrays of scalars. This conflates a minor formatting difference in a name with a wrong identifier, penalizes correctly extracted arrays that differ only in order, and draws no distinction between a missing field and an explicit null. Reliable evaluation requires the schema to define not just \emph{what} to extract but \emph{how} each field should be scored.
\end{itemize}
We introduce \textbf{ExtractBench}, an open-source benchmark and evaluation framework for PDF-to-JSON structured extraction. The dataset consists of (PDF, JSON Schema, gold JSON) triplets across 5 high-value domains, spanning schema complexity from tens to hundreds of fields. Additionally, we propose and open-source a \emph{schema-driven} evaluation framework: the schema specifies not only \emph{what} to extract but \emph{how} each field should be scored e.g., exact match for identifiers, numeric tolerance for quantities, semantic equivalence for free text.

Our key finding is that frontier models fail at enterprise-scale PDF to JSON tasks.
Evaluations of frontier models on ExtractBench reveal sharp degradation as schema breadth grows, reaching \textbf{0\% valid output} on a 369-field financial reporting schema across all tested frontier models. This exposes a failure mode hidden by benchmarks limited to small schemas:  frontier models become unreliable when required to generate long, deeply structured JSON. Valid JSON also does not imply correct extraction -- on one domain, models achieve 90\% valid output but only a 12.5\% pass rate.

Our paper makes four contributions:
\begin{itemize}[topsep=0pt, noitemsep]
    \item \textbf{ExtractBench dataset}: (PDF, JSON Schema, gold JSON) triplets with human-annotated gold labels across multiple domains and a wide range of schema complexities.
    \item \textbf{Schema-driven evaluation}: an extensible framework for nested structured extraction with per-field metrics, semantic array matching, and explicit missing/null handling.
    \item \textbf{Frontier baselines}: evaluations of frontier models from OpenAI, Anthropic, and Google under a consistent extraction setup.
    \item \textbf{Complexity analysis}: empirical characterization of failure modes as a function of schema depth, breadth, and array structure, identifying schema breadth as the dominant predictor of reliability.
\end{itemize}
Both the dataset and evaluation framework are released at \githubrepo{}.

\emph{Paper organization.}
Section~\ref{sec:related} reviews related benchmarks and structured output evaluation. Section~\ref{sec:dataset} introduces the ExtractBench dataset. Section~\ref{sec:evaluation} presents our schema-driven evaluation methodology. Section~\ref{sec:experiments} reports baseline results and analyzes performance as a function of schema complexity.
\subsection{Related Work}
\label{sec:related}

ExtractBench addresses a gap at the intersection of document understanding, structured output generation, and hierarchical JSON evaluation. We briefly review each area.

\paragraph{Document understanding and enterprise document IE benchmarks}
Document understanding benchmarks span form understanding, receipt parsing, and document VQA, including FUNSD~\cite{jaume2019funsd}, 
SROIE~\cite{huang2019icdar2019}, 
CORD~\cite{park2019cord}, 
and DocVQA~\cite{mathew2021docvqa}. 
Long-document settings and layout variability are emphasized by Kleister~\cite{stanislawek2021kleister} 
and visually-rich benchmarks such as ChartQA~\cite{masry2022chartqa} 
and InfographicVQA~\cite{mathew2022infographicvqa}.
Closest to our setting are enterprise-focused PDF benchmarks such as RealKIE~\cite{townsend2024realkie} 
and VRDU~\cite{wang2023vrdu}, which foreground OCR noise, sparse annotations, and hierarchical entities. However, these benchmarks typically evaluate span/entity extraction with F1-style metrics and comparatively small field sets (e.g., RealKIE: 3--28 fields; VRDU: $\sim$12 entities), rather than end-to-end extraction into large hierarchical JSON instances governed by JSON Schema. While these benchmarks address document understanding, they do not target schema-conformant JSON outputs.

\paragraph{Structured output generation and constrained decoding}
A parallel line of work studies generating schema-conformant structured outputs from language models, often via constrained decoding. JSONSchemaBench~\cite{geng2025generating} 
provides a large-scale evaluation of real-world JSON Schemas and shows that constrained decoding frameworks differ substantially in coverage and reliability.
Representative systems include Outlines~\cite{willard2023efficient}, 
XGrammar~\cite{dong2025xgrammar}, 
Guidance~\cite{guidance2024guidance}, 
and SGLang~\cite{zheng2024sglang}.
These efforts primarily target output validity and framework support for schema features; they do not address document-grounded extraction correctness or PDF-induced failure modes.
Adjacent evaluations in tool-use/function-calling, such as BFCL~\cite{patil2025the} 
and ToolLLM~\cite{qin2024toolllm}, similarly emphasize producing well-formed structured arguments, but operate over API signatures rather than document-derived schemas and gold labels. Orthogonally to document understanding, these benchmarks focus on validity rather than extraction correctness.

\paragraph{Nested JSON extraction and evaluation methodology}
Deep-JSON-Eval~\cite{zhou2025deepjsoneval} studies nested JSON extraction from long web text and reports degradation with increasing nesting, but uses text inputs and comparatively modest schema breadth.
Other benchmarks evaluate structured generation across formats (StructEval~\cite{yang2025structeval}) or focus on schema-valid JSON generation
 (json-mode-eval~\cite{nousresearch2024jsonmodeeval}), often relying on syntax or keyword-style checks.
In contrast, structured extraction from documents requires evaluation that is sensitive to field types (exact IDs vs.\ tolerant numbers vs.\ semantic strings), supports array alignment under reordering/omissions/spurious items, and distinguishes omission from hallucination via explicit missing/null semantics. ExtractBench is designed around these requirements via schema-driven evaluation, treating the schema as an executable specification of both structure and scoring.

\paragraph{PDF processing and upstream document parsing}
End-to-end document extraction depends on upstream PDF parsing and OCR. Recent systems such as Nougat~\cite{blecher2023nougat} and Docling~\cite{auer2024docling} improve PDF-to-text/markup extraction, while document representation and layout benchmarks/models (LayoutLM~\cite{xu2020layoutlm}, 
LayoutLMv3~\cite{huang2022layoutlmv3}, 
DocLayNet~\cite{pfitzmann2022doclaynet}, 
and TableFormer~\cite{nassar2022tableformer}) support robust layout and table understanding.
OCR and parsing errors can cascade into downstream IE degradation~\cite{van2020assessing}, motivating evaluation that measures the full pipeline rather than assuming clean text inputs. ExtractBench explicitly evaluates PDF-to-JSON extraction, capturing these upstream effects in benchmark outcomes.

\section{The ExtractBench Dataset}
\label{sec:dataset}

ExtractBench is a \emph{diagnostic benchmark} designed to reveal where and why frontier models fail on structured extraction. Rather than pursuing scale, we prioritize \textbf{complexity coverage}: 35 documents across 5 schemas spanning 2,076 pages and 12,867 evaluatable fields -- sufficient to expose systematic failure patterns while ensuring high-quality gold annotations through 67.9 hours of expert effort.

\subsection{Design Philosophy}
\label{subsec:philosophy}

ExtractBench follows a \textbf{quality over quantity} design. Each domain was selected to probe a distinct failure mode, and together they span orthogonal complexity dimensions that interact to reveal different model deficiencies:

\begin{itemize}
    \item \textbf{Schema breadth}: SEC 10-K/Q filings require 369 schema fields -- an order of magnitude larger than existing benchmarks. This alone causes \emph{complete failure}: 0\% valid output across all frontier models.
    \item \textbf{Output volume}: Research papers contain 100+ citation arrays, expanding to ${\sim}$25k output tokens despite only 16 schema fields. GPT models, which succeed on credit agreements (0.9k tokens), fail here.
    \item \textbf{Document length}: Credit agreements span 100--250 pages, exposing Claude's 100-page PDF limit while other models achieve 85\%+ accuracy.
    \item \textbf{Nesting depth}: Sports results reach depth 6 with nested athlete records, yet achieve 90\% validity -- demonstrating that depth alone is a weak predictor of failure.
\end{itemize}

This deliberate curation ensures ExtractBench tests failure modes that arise in production, rather than sampling documents at random. The benchmark reveals that schema breadth and output volume -- not document length or nesting depth -- are the dominant predictors of extraction failure.

\begin{figure*}[ht]
    \centering
    \begin{tikzpicture}

    \begin{axis}[
        name=tokenplot,
        ybar,
        width=7.5cm,
        height=5cm,
        bar width=10pt,
        ymode=log,
        log origin=infty,
        ylabel={Avg.\ Tokens (k)},
        ylabel style={font=\small},
        title={\small Input vs Output Tokens},
        symbolic x coords={10-K/Q, Credit, Research, Resumes, Sports},
        xtick=data,
        x tick label style={rotate=25, anchor=east, font=\footnotesize},
        ymin=0.3, ymax=500,
        ytick={1,10,100},
        yticklabels={1k,10k,100k},
        y tick label style={font=\footnotesize},
        legend style={
            anchor=north east,
            font=\footnotesize,
        },
        enlarge x limits=0.12,
    ]
    \addplot[fill=inputpdf, draw=inputpdf!80!black] coordinates {
        (10-K/Q, 139.6)
        (Credit, 335.5)
        (Research, 104.3)
        (Resumes, 7.6)
        (Sports, 7.0)
    };
    \addplot[fill=outputjson, draw=outputjson!80!black] coordinates {
        (10-K/Q, 2.9)
        (Credit, 0.52)
        (Research, 23.0)
        (Resumes, 1.2)
        (Sports, 0.38)
    };
    \legend{Input (Vision), Output (JSON)}
    \end{axis}

    \begin{axis}[
        at={(tokenplot.east)},
        anchor=west,
        xshift=1.7cm,
        xbar,
        width=6cm,
        height=5cm,
        bar width=10pt,
        xlabel={Compression Ratio},
        xlabel style={font=\small},
        title={\small Input/Output Ratio},
        ytick=data,
        yticklabels={10-K/Q, Credit, Research, Resumes, Sports},
        y tick label style={font=\footnotesize},
        x tick label style={font=\footnotesize},
        y dir=reverse,
        xmin=0, xmax=750,
        xtick={0,200,400,600},
        enlarge y limits=0.12,
    ]
    \addplot[fill=comprgreen, draw=comprgreen!80!black] coordinates {
        (49,0)
        (682,1)
        (10.3,2)
        (6.0,3)
        (19.7,4)
    };
    \end{axis}

    \end{tikzpicture}
    \caption{Token statistics by domain (Vision tokens). Left: average input and output token counts on log scale. Right: compression ratio (input$\div$output). Credit agreements show 682$\times$ compression (needle-in-haystack extraction from 137-page average documents), while research papers show only 10$\times$ despite producing the highest output volume (23k tokens).}
    \label{fig:token-compression}
\end{figure*}
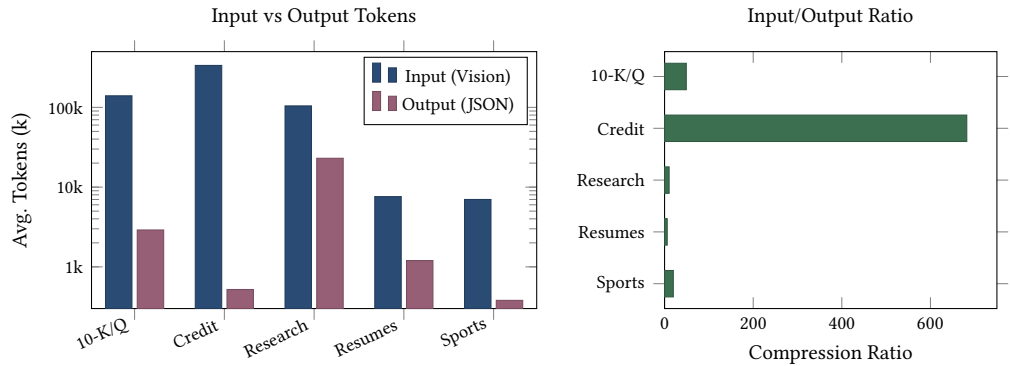


Figure~\ref{fig:token-compression} reveals two contrasting task types: credit agreements are ``needle-in-haystack'' tasks (682$\times$ compression) requiring models to locate sparse fields in lengthy legal prose, while research papers are ``enumeration'' tasks with the lowest compression ratio (10$\times$) despite producing the highest output volume (23k tokens). This explains why credit agreements -- despite being the longest documents -- achieve the highest pass rate: \emph{output volume, not input length, is the binding constraint}. More details are presented in Table~\ref{tab:token-stats} of Appendix \ref{app:tables-and-figures}.

\subsection{Domains and Complexity Dimensions}
\label{subsec:domains}

We selected five domains from economically valuable settings, each presenting distinct extraction challenges:

\begin{table}[ht]
\centering
\caption{Domain characteristics and complexity dimensions. Domains are ordered by schema breadth (number of evaluatable fields).}
\label{tab:domain-complexity}
\footnotesize
\begin{tabular}{lrrrl}
\toprule
\textbf{Domain} & \textbf{Keys} & \textbf{Depth} & \textbf{Docs} & \textbf{Primary Challenge} \\
\midrule
Sports Results & 12 & 6 & 5 & Deep nesting, tabular hierarchies \\
Credit Agreements & 13 & 3 & 10 & Long documents (100--250 pages) \\
Research Papers & 16 & 5 & 6 & Long arrays (100+ citations) \\
Resumes & 31 & 4 & 7 & Semi-structured layouts \\
SEC 10-K/Qs & 369 & 4 & 7 & Enterprise-scale schema breadth \\
\bottomrule
\end{tabular}
\end{table}

\paragraph{Schema complexity dimensions.} We characterize complexity along three axes: \emph{breadth} (number of schema fields), \emph{depth} (maximum nesting level), and \emph{array complexity} (presence and length of repeated structures). Crucially, these dimensions are orthogonal and interact: Sports has depth 6 but achieves 90\% validity because its tabular structure is repetitive; Research has only depth 5 but achieves 39\% validity because 100+ citation items overwhelm output capacity. ExtractBench is the first benchmark to systematically vary these dimensions, enabling diagnosis of which factors drive failure.

\begin{table}[ht]
\centering
\caption{Dataset scale by domain. Gold values count all leaf values including array item expansion.}
\label{tab:objects}
\begin{tabular}{lrrr}
\toprule
\textbf{Domain} & \textbf{Documents} & \textbf{Pages} & \textbf{Gold Values} \\
\midrule
SEC 10-K/Qs & 7 & 422 & 9,071 \\
Professional Resumes & 7 & 21 & 1,007 \\
Credit Agreements & 10 & 1,368 & 269 \\
Sports Results & 5 & 15 & 522 \\
Research Papers & 6 & 250 & 1,998 \\
\midrule
\textbf{Total} & \textbf{35} & \textbf{2,076} & \textbf{12,867} \\
\bottomrule
\end{tabular}
\end{table}

\subsection{Annotation Process}
\label{subsec:annotation}

All gold extractions were created through human annotation to ensure reliability. Annotators received the source PDF, target JSON schema, and detailed guidelines for ambiguous cases. Each annotation underwent validation for schema conformance and extraction accuracy.
Annotation hours breakdown are shown in Table \ref{tab:annotation-hours}. The 67.9 hours of annotation effort -- with SEC 10-K/Qs alone requiring 56 hours -- reflects the investment needed for reliable gold standards on complex schemas. 

\begin{table}[ht]
\centering
\caption{Human annotation effort. SEC 10-K/Qs required 8 hours per document due to schema complexity (369 fields).}
\label{tab:annotation-hours}
\begin{tabular}{llrr}
\toprule
\textbf{Domain} & \textbf{Hours} & \textbf{Hrs/Doc} \\
\midrule
SEC 10-K/Qs  & 56.1 & 8.0 \\
Professional Resumes  & 5.0 & 0.7 \\
Credit Agreements  & 2.9 & 0.3 \\
Sports Results  & 2.5 & 0.5 \\
Research Papers  & 1.4 & 0.2 \\
\midrule
\textbf{Aggregate} &  \textbf{67.9} & 2.7 \\
\bottomrule
\end{tabular}
\end{table}

\section{A Principled Evaluation Methodology}
\label{sec:evaluation}

Structured extraction demands evaluation beyond a single global metric. Realistic schemas mix identifiers, numbers, text, optional fields, and nested arrays -- each with different correctness criteria. We present a schema-driven evaluation framework, a contribution independent of the dataset, that functions as \emph{infrastructure for the field}: extensible, principled, and reusable beyond ExtractBench.

\subsection{The Evaluation Challenge}
\label{subsec:challenge}

Consider a credit agreement schema with heterogeneous requirements: an \emph{agreement ID} demands exact match; a \emph{principal amount} requires numeric tolerance (rounding vs.\ error); a \emph{borrower name} needs semantic equivalence (``ABC Corp'' $\approx$ ``ABC Corporation''); \emph{covenants} form arrays of complex objects where order is irrelevant.

A single global metric cannot capture these semantics. Exact match over-penalizes benign variation; fuzzy similarity cannot distinguish critical identifiers from descriptive text; embedding similarity is poorly defined for numeric precision. Nested schemas amplify the problem: enterprise schemas (e.g., a 10-K/Q with 369 fields) require recursive evaluation while preserving per-field semantics and differentiating structural failures from content errors.

Evaluation must also distinguish error modes: \emph{structural} errors (invalid JSON, wrong nesting), \emph{omission} (missing a value that exists), \emph{hallucination} (producing a value absent from the source), and \emph{accuracy} errors (extracting a value incorrectly). Our insight: \textbf{the schema should define not just \emph{what} to extract, but \emph{how} to evaluate each field}. This principle -- schema-driven evaluation -- is the foundation of our methodology.

\subsection{Schema-Driven Evaluation}
\label{subsec:architecture}

We treat the JSON Schema as an executable specification via \emph{AST-based dual-traversal}: the schema is parsed into a typed abstract syntax tree (AST), then traversed jointly with the gold and predicted JSON instances. Each node represents a schema construct (object, array, primitive, \texttt{anyOf}, \texttt{\$ref}) and carries an evaluation configuration specifying which metric to apply. Evaluation is local -- each node computes a result from its configuration and the paired values -- and global scores are aggregations over the tree.

This architecture provides key properties for scalability and extensibility: (1) recursive traversal handles arbitrarily nested schemas including \texttt{\$ref} resolution; (2) independent fields evaluate in parallel, essential for 369-field schemas where sequential evaluation would be prohibitively slow; and (3) the visitor pattern separates traversal logic from metric computation, so new metrics integrate via a plugin registry without modifying core code.

\paragraph{Metric library.} We provide a library of metrics (Table~\ref{tab:presets}) covering common patterns: exact and case-insensitive string matching, fuzzy matching via Levenshtein distance, LLM-based semantic equivalence for free text, exact and tolerance-based numeric comparison, and semantic array alignment. Each field declares its metric via an \texttt{evaluation\_config} annotation:

{\footnotesize
\begin{verbatim}
"borrower_name": {"type": "string",
                  "evaluation_config": "string_semantic"}
"principal":     {"type": "number",
                  "evaluation_config": {"metric_id": "number_tolerance",
                                        "params": {"tolerance": 0.001}}}
\end{verbatim}
}

For LLM-based metrics, optional \texttt{additional\_instructions} encode domain equivalences (e.g., treating ``Inc'' and ``Incorporated'' as equivalent), letting domain experts specify correctness criteria without modifying code. New metrics are added via a plugin registry without modifying core evaluation logic.

\begin{table}[t]
\centering
\caption{Evaluation presets. Each maps to a metric with default parameters.}
\small
\label{tab:presets}
\begin{tabular}{lll}
\toprule
\textbf{Preset} & \textbf{Metric} & \textbf{Use Case} \\
\midrule
\texttt{string\_exact} & Exact match & IDs, codes \\
\texttt{string\_case\_insensitive} & Case-folded & Enums, status \\
\texttt{string\_fuzzy} & Levenshtein & Typo tolerance \\
\texttt{string\_semantic} & LLM equivalence & Free text \\
\texttt{number\_exact} & Exact numeric & Counts \\
\texttt{number\_tolerance} & Within margin & Amounts (0.1\%) \\
\texttt{boolean\_exact} & Exact boolean & Flags \\
\texttt{array\_llm} & Semantic alignment & Object arrays \\
\bottomrule
\end{tabular}
\end{table}

\subsection{Missing Value Semantics}
\label{subsec:missing}

We distinguish three value states: \emph{present} (field has extracted data), \emph{null} (field is explicitly empty), and \emph{MISSING} (field absent from output). This three-way distinction enables precise error categorization. When gold has a value but predicted is MISSING, the error is \emph{omission} -- the model failed to extract existing information. When gold is \textit{null} but predicted has a value, the error is \emph{hallucination} -- the model fabricated information not in the source. These failure modes have different downstream risk profiles: hallucination may introduce false information into pipelines, while omission loses true information. Separating them is essential for diagnosis and for understanding model behavior under different schema configurations. All metrics implement a policy-aware interface that resolves the (gold, predicted) state combination before comparing values, ensuring uniform handling across metric types.

\subsection{Semantic Array Matching}
\label{subsec:arrays}

Arrays of objects require alignment before scoring: which predicted item corresponds to which gold item? Position-based matching fails under reordering, omissions, and spurious items.

We use LLM-based semantic alignment. The judge receives the item schema, gold items, predicted items, and per-field evaluation criteria, then returns a mapping identifying \emph{matched pairs}, \emph{missed gold items} (false negatives), and \emph{spurious predictions} (false positives, potential hallucinations). From these sets we compute precision, recall, and F1 at the array level. For matched pairs, we recursively evaluate item contents using the item schema's field-level metrics, propagating the schema-driven approach into nested structures.

The matcher handles edge cases including empty arrays, partial matches (aligned on key fields with differences captured in per-field scores), duplicate predictions, and large arrays processed in batches to respect context limits. Our evaluation infrastructure allows extension of more sophisticated matching strategies. 

\subsection{Open-Source Release}
\label{subsec:tool}

We release the evaluation framework as open-source infrastructure alongside ExtractBench at \githubrepo{} under permissive MIT license. The framework is designed for extension: new metrics register via a plugin interface without modifying core traversal logic; domain experts encode evaluation rules directly in JSON Schema annotations rather than writing code; and alternative array matchers can be substituted for domains requiring specialized alignment. This positions the framework as reusable infrastructure -- not a one-off scoring script -- enabling the community to benchmark extraction systems under a consistent, extensible methodology.

\begin{table*}[t]
    \centering
    \small
    \caption{Baseline performance on ExtractBench. \emph{Valid JSON}: fraction producing parseable, schema-conforming JSON. \emph{Pass Rate}: field-level pass rate (invalid extractions contribute 0). \emph{Acc (Valid)}: pass rate conditioned on valid extraction only. Domain subheaders show (documents / avg.\ pages: \textcolor{purple}{schema keys} / \textcolor{gold}{avg.\ gold tokens}). Best per column in \textbf{bold}; complete failure in \textcolor{red}{red}.}
    \label{tab:main-results}
    {%
    \begin{tabular}{lcccccccc}
    \toprule
    & & \multicolumn{5}{c}{\textbf{Domain Pass Rate (Num Docs/Avg Pages: Num Keys / Avg Gold Tokens)}} & & \\
    \cmidrule(lr){3-7}
    \textbf{Model} & \textbf{Valid} & \textbf{Credit} & \textbf{Research} & \textbf{Resumes} & \textbf{Sports} & \textbf{SEC 10-K/Q} & \textbf{Overall} & \textbf{Acc} \\
    & \textbf{JSON} & \scriptsize(10docs/137pgs: \textcolor{purple}{13}/\textcolor{gold}{0.9k}) & \scriptsize(6docs/42pgs: \textcolor{purple}{16}/\textcolor{gold}{25k}) & \scriptsize(7docs/3pgs: \textcolor{purple}{31}/\textcolor{gold}{3k}) & \scriptsize(5docs/3pgs: \textcolor{purple}{12}/\textcolor{gold}{3k}) & \scriptsize(7docs/60pgs: \textcolor{purple}{369}/\textcolor{gold}{24k}) & \textbf{Pass Rate} & \textbf{Valid} \\
    \midrule
    Gemini 3 Flash & \textbf{25/35} & 111/130 (85.4\%) & \textbf{46/96 (47.9\%)} & 45/217 (20.7\%) & \textbf{11/60 (18.3\%)} & \textcolor{red}{0/2583 (0.0\%)} & \textbf{213/3086 (6.9\%)} & 74.7\% \\
    Gemini 3 Pro & 21/35 & 110/130 (84.6\%) & 5/96 (5.2\%) & 46/217 (21.2\%) & 8/60 (13.3\%) & \textcolor{red}{0/2583 (0.0\%)} & 169/3086 (5.5\%) & 74.8\% \\
    GPT-5.2 & 21/35 & 105/130 (80.8\%) & 6/96 (6.2\%) & 43/217 (19.8\%) & 8/60 (13.3\%) & \textcolor{red}{0/2583 (0.0\%)} & 162/3086 (5.2\%) & 71.7\% \\
    GPT-5 & 13/35 & \textbf{113/130 (86.9\%)} & \textcolor{red}{0/96 (0.0\%)} & 8/217 (3.7\%) & 2/60 (3.3\%) & \textcolor{red}{0/2583 (0.0\%)} & 123/3086 (4.0\%) & \textbf{80.4\%} \\
    Sonnet 4.5 & 15/35 & \textcolor{red}{0/130 (0.0\%)} & 47/96 (49.0\%) & 46/217 (21.2\%) & 8/60 (13.3\%) & \textcolor{red}{0/2583 (0.0\%)} & 101/3086 (3.3\%) & 66.9\% \\
    Opus 4.5 & 12/35 & \textcolor{red}{0/130 (0.0\%)} & 16/96 (16.7\%) & \textbf{52/217 (24.0\%)} & 8/60 (13.3\%) & \textcolor{red}{0/2583 (0.0\%)} & 76/3086 (2.5\%) & 65.0\% \\
    \midrule
    \textbf{Aggregate} & 107/210 & 439/780 (56.3\%) & 120/576 (20.8\%) & 240/1302 (18.4\%) & 45/360 (12.5\%) & \textcolor{red}{0/15498 (0.0\%)} & 844/18516 (4.6\%) & 72.9\% \\
    \bottomrule
    \end{tabular}%
    }
\end{table*}

\section{Experiments}
\label{sec:experiments}

We evaluate six frontier LLMs on ExtractBench to establish baselines and characterize performance across complexity dimensions:

\begin{itemize}
    \item \textbf{Google Gemini}: Gemini 3 Pro, Gemini 3 Flash
    \item \textbf{OpenAI GPT}: GPT-5.2, GPT-5
    \item \textbf{Anthropic Claude}: Claude Sonnet 4.5, Claude Opus 4.5
\end{itemize}

\textbf{Setup.} All models receive the source PDF via their provider's native multimodal API, the target JSON schema, and a zero-shot extraction prompt. Default API parameters are used throughout. Extracted outputs are evaluated against gold annotations using the schema-driven framework described in \S\ref{sec:evaluation}, with Gemini 2.5 Flash as the LLM judge (pass threshold 0.7) and a Levenshtein similarity threshold of 0.8 for fuzzy matching.

We report two complementary metrics. \emph{Valid JSON} measures the fraction of extractions producing parseable, schema-conforming JSON -- a necessary but insufficient condition for correctness. \emph{Pass rate} measures content accuracy: each schema field is scored against the gold annotation using the metric declared in the schema's \texttt{evaluation\_config} (e.g., exact match for identifiers, semantic equivalence for free text; see Table~\ref{tab:presets}), and a field passes if its score meets the metric-specific threshold. Invalid extractions contribute zero passed fields but still count in the denominator, so pass rate reflects end-to-end reliability.

\subsection{Main Results}
\label{subsec:results}

Table~\ref{tab:main-results} presents the full results. Across 210 extraction attempts, the six models achieve a 51\% valid JSON rate (107/210), with an aggregate pass rate of 4.6\% (844/18,516 field evaluations across all models).\footnote{Each of the 35 documents contributes one evaluation per schema field per model; 3,086 field positions $\times$ 6 models = 18,516 total evaluations.} The best-performing model, Gemini 3 Flash, reaches only 6.9\%. The headline finding is the complete failure on SEC 10-K/Q filings: \textit{no model produced valid output for any of the seven documents}. This domain's 369-field schema accounts for 84\% of all field evaluations (15,498/18,516), dominating the aggregate. Even excluding 10-K/Q, the pass rate is 28.0\% (844/3,018 fields).

\textbf{Valid extractions achieve reasonable accuracy.}
The rightmost column of Table~\ref{tab:main-results} shows that when models produce valid JSON, field-level accuracy ranges from 65\% to 80\%, with an aggregate of 72.9\% across all valid extractions. This contrasts sharply with the 4.6\% end-to-end pass rate. However, interpreting this gap requires caution: valid-only accuracy is computed on a \emph{biased sample} -- the extractions that succeeded tend to be the easier cases (simpler schemas, shorter outputs). We cannot assume models would achieve similar accuracy on the hard cases if they merely fixed formatting. The results suggest that frontier LLMs possess reasonable extraction capability on tractable tasks, but this capability degrades in ways that are entangled with the same complexity factors that cause format failures.

\textbf{Model patterns.}
Gemini models achieve the strongest overall results, with Flash leading on both validity (71\%, 25/35) and pass rate (6.9\%). Notably, Flash outperforms the larger Pro model across most domains. GPT-5 has the lowest validity (37\%, 13/35) but the highest valid-only accuracy (80.4\%); however, this likely reflects selection bias -- GPT-5 succeeds primarily on the easiest domain (credit agreements), where all models achieve high accuracy. Claude models achieve competitive extraction accuracy on research papers and resumes but are systematically blocked on credit agreements by Anthropic's 100-page PDF ingestion limit\footnote{\url{https://platform.claude.com/docs/en/build-with-claude/pdf-support}}, producing 0/130 fields in that domain.

\textbf{Domain patterns.}
Credit agreements -- the simplest schema at 13 fields -- achieve the highest pass rate (56.3\%) across models that can ingest the documents. At the opposite extreme, 10-K/Q filings (369 fields) yield 0\% valid output. Between these poles, an important dissociation emerges: valid JSON does not imply correct extraction. Sports results achieve the highest validity (90\%, 27/30) yet only a 12.5\% pass rate, because array-level evaluation penalizes structural discrepancies in deeply nested athlete records even when individual field values are largely correct. Research papers exhibit the sharpest model variation, with validity ranging from 0\% (GPT-5) to 83\% (Claude Sonnet 4.5, Gemini 3 Flash). Per-domain breakdowns by model appear in Appendix~\ref{app:tables-and-figures}.

\subsection{Analysis: Why and How Models Fail}
\label{subsec:analysis}

We analyze the complexity factors that drive extraction failure and characterize the specific failure modes that result.

\begin{table}[ht]
    \centering
    \small
    \caption{Per-document task complexity by domain, ordered by schema key count. \emph{Keys}: evaluatable schema fields. \emph{Depth}: maximum nesting level. Remaining columns are per-document averages. \emph{Tokens}: GPT-2 token count of gold JSON. \emph{Fields} and \emph{Arr.\ Items} decompose the extraction target into non-array fields and array items in gold annotations.}
    \label{tab:complexity}
    \begin{tabular}{lrrrrrr}
    \toprule
    \textbf{Domain} & \textbf{Pages} & \textbf{Keys} & \textbf{Depth} & \textbf{Tokens} & \textbf{Fields} & \textbf{Arr.\ Items} \\
    \midrule
    Sports & 3 & 12 & 6 & 3,416 & 4.0 & 19.8 \\
    Credit & 137 & 13 & 3 & 883 & 11.2 & 15.7 \\
    Research & 42 & 16 & 5 & 25,366 & 10.3 & 309.5 \\
    Resumes & 3 & 31 & 4 & 3,222 & 8.9 & 44.3 \\
    10-K/Qs & 60 & 369 & 4 & 24,418 & 317.0 & 182.3 \\
    \bottomrule
    \end{tabular}
\end{table}

\textbf{Schema breadth is the dominant predictor.}
Table~\ref{tab:complexity} summarizes per-document complexity by domain, ordered by schema key count. The trend is clear: credit agreements (13 keys) achieve the highest pass rate at 56.3\%, while 10-K/Q filings (369 keys) produce zero valid outputs across all models. The 10-K/Q schema requires generating ${\sim}$24,400 tokens of structured JSON per document, exceeding the reliable output capacity of every evaluated model.

\textbf{Output volume drives GPT failures.}
Gold JSON token count -- a proxy for required output complexity -- reveals a provider-specific pattern. GPT models succeed on credit agreements (0.9k gold tokens) but struggle on all other domains, which require 3k--25k tokens. GPT-5 achieves 100\% validity on credit agreements yet only 0--14\% elsewhere, suggesting that output volume is a binding constraint for this model family.

\textbf{Page count constrains Claude models.}
Claude's 100-page PDF ingestion limit blocks all credit agreement documents (97--218 pages, averaging 137). This eliminates Claude from the domain where other models perform best, illustrating how provider-specific API constraints interact with document properties to shape benchmark outcomes.

\textbf{Array complexity amplifies difficulty beyond schema size.}
Research papers have only 16 schema keys -- comparable to credit agreements (13) -- yet achieve far lower validity (39\% vs.\ 67\%). The difference lies in array complexity: research paper gold annotations average 309.5 array items per document (driven by citation lists of 100+ items), yielding 25,366 gold tokens despite a compact schema. Credit agreements average only 15.7 array items and 883 gold tokens. This demonstrates that \emph{total output volume} -- reflecting both schema fields and array expansion -- is a better predictor of difficulty than schema key count alone.

\textbf{Nesting depth is a weak predictor.}
Sports results have the deepest schema (depth 6) yet achieve the highest validity (90\%), while research papers (depth 5) achieve only 39\%. Deeply nested but repetitive tabular structures (e.g., age-group hierarchies) are structurally predictable, whereas deeply nested heterogeneous schemas (e.g., variable-field citation metadata) are harder to populate correctly. This suggests depth affects extraction accuracy more than format validity.

\textbf{Benchmark diversity.}
Resumes and sports results illustrate the orthogonality of ExtractBench's complexity dimensions. Both involve short documents ($\sim$3 pages, $\sim$3k gold tokens), but differ in schema breadth (31 vs.\ 12 keys) and array complexity (44.3 vs.\ 19.8 items per document). They exhibit distinct performance profiles: resumes achieve 62\% validity with 18.4\% pass rate, while sports achieve 90\% validity but only 12.5\% pass rate. ExtractBench's domains thus probe distinct failure modes rather than a single difficulty axis. 

\textbf{Failure mode breakdown.}
These complexity factors manifest as provider-specific failure modes. Of the 103 invalid extractions (49\% of all attempts), Table~\ref{tab:error-analysis} categorizes five types.

\begin{table}[ht]
    \centering
    \small
    \caption{Distribution of extraction failure modes across models. \emph{Trailing comma} and \emph{Truncated JSON} are malformed JSON errors. \emph{Empty response} indicates no JSON was returned. \emph{PDF page limit} and \emph{Context length} are API-level rejections.}
    \label{tab:error-analysis}
    \resizebox{\columnwidth}{!}{%
    \begin{tabular}{lrrrrrrr}
    \toprule
    \textbf{Error Type} & \textbf{Opus} & \textbf{Sonnet} & \textbf{GPT-5.2} & \textbf{GPT-5} & \textbf{G3 Pro} & \textbf{G3 Flash} \\
    \midrule
    Empty response & --- & --- & 14 & 22 & 5 & --- \\
    Trailing comma & 11 & 8 & --- & --- & 5 & 7 \\
    PDF page limit & 10 & 10 & --- & --- & --- & --- \\
    Truncated JSON & 1 & 1 & --- & --- & 4 & 3 \\
    Context length & 1 & 1 & --- & --- & --- & --- \\
    \midrule
    \textbf{Total failures} & \textbf{23} & \textbf{20} & \textbf{14} & \textbf{22} & \textbf{14} & \textbf{10} \\
    \bottomrule
    \end{tabular}%
    }
    \end{table}

\begin{itemize}
\item \emph{Empty response} (41 cases, 40\% of failures): the model returns no parseable output. This affects only GPT and Gemini 3 Pro; GPT-5 accounts for 22 cases. Nearly half (19/41) occur on 10-K/Q documents, suggesting these models silently fail when extraction complexity exceeds a threshold.
\item \emph{Trailing commas} (31 cases, 30\%): the model produces syntactically plausible JSON with trailing commas before closing braces -- valid in JavaScript but rejected by strict JSON parsers. This error is exclusive to Claude and Gemini, and concentrated in the 10-K/Q domain (20/31), where the 369-field schema exceeds the models' ability to maintain syntactic discipline over long outputs.
\item \emph{PDF page limit} (20 cases, 19\%): Claude's 100-page limit causes rejection of all credit agreement documents for both Claude models.
\item \emph{Truncated JSON} (9 cases) and \emph{context length} errors (2 cases) account for the remaining failures, where models exhaust output or input token budgets.
\end{itemize}
Notably, nearly a third of all failures (trailing commas and truncation) stem from formatting issues that could potentially be addressed through post-processing or constrained decoding -- motivating the structured output experiment in \S\ref{subsec:structured-output}.

\subsection{Structured Output Mode}
\label{subsec:structured-output}

All three providers offer a \emph{structured output} mode--OpenAI's \emph{Structured Outputs}\footnote{\url{https://platform.openai.com/docs/guides/structured-outputs}}, Google's \emph{response schema}\footnote{\url{https://ai.google.dev/gemini-api/docs/structured-output}}, and Anthropic's \emph{JSON outputs}\footnote{\url{https://platform.claude.com/docs/en/build-with-claude/structured-outputs}}--in which the target JSON schema is supplied as a response-format constraint rather than embedded in the prompt. These features rely on \emph{constrained decoding}: the provider compiles the JSON schema into a grammar artifact (e.g., a finite-state automaton) that restricts the set of valid tokens at each generation step, guaranteeing syntactically valid, schema-conforming output. OpenAI's documentation notes that ``the first request you make with any schema will have additional latency as our API processes the schema,'' consistent with upfront grammar compilation; Anthropic describes this as ``constrained sampling with compiled grammar artifacts.'' In principle, constrained decoding eliminates the trailing-comma and truncated-JSON failures that accounted for 39\% of errors in our prompt-mode experiments (\S\ref{subsec:analysis}).

We repeated the full 210-extraction experiment using each provider's structured output API. The results (Table~\ref{tab:structured-model}) show that structured outputs \emph{reduce} both validity and accuracy relative to prompt-based extraction: overall validity dropped from 51\% (107/210) to 37\% (77/210), and the best-model pass rate fell from 6.9\% to 5.5\%.

\begin{table}[ht]
    \centering
    \caption{Field-level pass rate by model and domain with structured outputs. Best per domain in \textbf{bold}. SEC 10-K/Qs and Resumes omitted (0\% validity across all models). Invalid extractions contribute 0 for all fields.}
    \label{tab:structured-model}
    \begin{tabular}{lcccc}
    \toprule
    \textbf{Model} & \textbf{Credit} & \textbf{Research} & \textbf{Sports} & \textbf{Overall} \\
    \midrule
    Gemini 3 Pro & \textbf{88.5\%} & 9.4\% & 6.7\% & 4.1\% \\
    Gemini 3 Flash & 83.8\% & \textbf{54.2\%} & 13.3\% & \textbf{5.5\%} \\
    GPT-5.2 & 83.1\% & 51.0\% & \textbf{16.7\%} & 5.4\% \\
    GPT-5 & 70.0\% & 32.3\% & 3.3\% & 4.0\% \\
    Claude Opus 4.5 & 0.0\% & 0.0\% & \textbf{16.7\%} & 0.3\% \\
    Claude Sonnet 4.5 & 0.0\% & 0.0\% & \textbf{16.7\%} & 0.3\% \\
    \midrule
    \textbf{Total} & 54.2\% & 14.7\% & 7.3\% & 3.0\% \\
    \bottomrule
    \end{tabular}
    \end{table}

\textbf{Schema rejection.} Each provider's constrained decoding mode supports only a \emph{subset} of JSON Schema. Anthropic prohibits recursive schemas and rejects schemas exceeding internal complexity thresholds; Google's API ``may reject very large or deeply nested schemas.'' These restrictions create a hard failure mode absent from prompt-based extraction, where the schema is merely instructional text. In our experiments, the resume schema -- which produced 62\% validity in prompt mode (26/42) -- was rejected outright by all providers in structured mode (0/42). Claude models dropped from 12--15/35 valid extractions to just 5/35, succeeding only on sports statistics schemas.

\textbf{Accuracy degradation.} Even when structured mode accepts a schema and produces valid JSON, extraction accuracy can decline. GPT-5's pass rate on credit agreements fell from 86.9\% to 70.0\%, suggesting that the constrained decoding overhead -- maintaining a valid grammar state across thousands of tokens -- may compete with the model's capacity to attend to document content. This effect is most pronounced on long-form extractions where the model must simultaneously satisfy structural constraints and resolve complex cross-references in the source document.

Credit agreements remained the strongest domain at 54.2\% pass rate (423/780), comparable to the 56.3\% achieved in prompt mode. SEC 10-K/Q filings remained at 0\% validity, confirming that the 369-field schema exceeds structured output capabilities as well.

\textbf{Why constrained decoding may underperform on complex schemas.}
The paradoxical result -- that a mechanism designed to guarantee valid JSON actually \emph{lowers} validity -- likely stems from a mismatch between the assumptions of structured output APIs and the demands of complex extraction schemas. Two factors may contribute:
(1) \textbf{Grammar complexity scaling.} The compiled grammar grows with the number of schema properties, nesting depth, and array cardinality. For schemas with hundreds of fields (e.g., our 369-key 10-K/Q schema), the resulting automaton may exceed provider-internal limits, or the per-token constraint-checking overhead may degrade generation quality by effectively reducing the model's usable capacity for content reasoning.
(2) \textbf{Rigid structural enforcement.} In prompt mode, a model that encounters difficulty with one section of a schema can skip fields, emit null values, or restructure its output -- preserving partial correctness. Constrained decoding enforces the full schema structure at every token, leaving no room for graceful degradation. If the model begins populating a deeply nested array and runs low on output budget, it cannot truncate cleanly; the grammar requires all mandatory fields to be emitted, potentially forcing the generation into a degenerate state or exceeding token limits.

These findings indicate that structured output modes, while effective for simple, shallow schemas, introduce additional failure modes for complex extraction tasks. Constrained decoding does not yet serve as a reliable substitute for prompt-based schema specification when schemas are large, deeply nested, or use JSON Schema features outside each provider's supported subset.

\section{Conclusion}
\label{sec:conclusion}

We introduced ExtractBench, a benchmark and evaluation framework for PDF-to-JSON structured extraction. ExtractBench makes two contributions: (1) a dataset of (PDF, JSON Schema, gold JSON) triplets spanning schema complexities from 13 to 369 fields, and (2) a schema-driven evaluation methodology that captures per-field semantics, distinguishes omission from hallucination, and handles nested arrays via semantic alignment.
Baseline evaluations reveal that frontier models remain unreliable on realistic enterprise schemas, with performance degrading sharply as schema breadth increases—culminating in 0\% valid output on 369-field financial reports. Our analysis shows that \emph{output volume}, not input length or nesting depth, is the dominant predictor of failure. This exposes a gap between current capabilities and production requirements for complex structured extraction.

Both the dataset and evaluation framework are released at \githubrepo{} under MIT license. ExtractBench enables several research directions such as the impact of constrained decoding on large schemas and decomposition strategies for handling complex schemas. Long complex documents require exploration of context-management approaches like retrieval and agentic methods. We hope ExtractBench provides the infrastructure for the field to make targeted progress on further scaling AI capabilities.





\bibliographystyle{ACM-Reference-Format}
\bibliography{references}


\appendix
\section{Ethical Considerations}
\label{sec:ethics}


\paragraph{Data Privacy and Licensing}
\label{subsec:privacy}
All documents in ExtractBench are from public domain sources or synthetically generated:

\begin{itemize}
    \item \textbf{SEC 10-K/Q filings}: Public domain, sourced from the SEC EDGAR database.
    \item \textbf{Credit agreements}: Public domain, sourced from SEC EDGAR filings (exhibit attachments to 10-K/8-K filings).
    \item \textbf{Research papers}: Public domain or open-access publications.
    \item \textbf{Sports results}: Public domain, sourced from publicly available competition result postings.
    \item \textbf{Professional resumes}: Synthetically generated. No real personally identifiable information (PII) is present; all names, contact details, employment histories, and educational backgrounds are fictional.
\end{itemize}

No proprietary or restricted-access documents are included.

\paragraph{Limitations}
\label{subsec:limitations}
We acknowledge several limitations of ExtractBench:
\begin{itemize}
    \item \textbf{Language}: All documents are in English.
    \item \textbf{Geographic scope}: Financial documents (10-K/Qs, credit agreements) are US-centric, reflecting SEC filing requirements.
    \item \textbf{Domain coverage}: The dataset spans five domains (SEC filings, credit agreements, research papers, resumes, sports) only.
    \item \textbf{Demographic diversity}: Synthetic resumes may not fully represent the diversity of real-world professional backgrounds.
    \item \textbf{Dataset size}: With 35 documents, the dataset is designed for quality diagnostic evaluation rather than quantity statistical power across fine-grained subgroups.
    \item \textbf{Annotation team}: Gold annotations were created by a team of three annotators. We do not report inter-annotator agreement statistics; instead, all annotations underwent review for schema conformance and extraction accuracy.
\end{itemize}

We do not anticipate significant misuse potential, as the benchmark focuses on extraction accuracy measurement rather than generating sensitive content.

\section{Additional Tables and Figures}\label{app:tables-and-figures}

Tables~\ref{tab:domain-10kq}--\ref{tab:domain-sports} provide per-domain breakdowns of model performance, complementing the aggregate results in Table~\ref{tab:main-results}. Table~\ref{tab:token-stats} provides per-document token statistics.

\begin{table}[ht]
\centering
\small
\caption{Model performance on SEC 10-K/Q filings (7~documents).}
\label{tab:domain-10kq}
\begin{tabular}{lcc}
\toprule
\textbf{Model} & \textbf{Valid} & \textbf{Pass Rate} \\
\midrule
Claude Opus 4.5 & 0/7 & 0/2583 (0\%) \\
Claude Sonnet 4.5 & 0/7 & 0/2583 (0\%) \\
Gemini 3 Flash & 0/7 & 0/2583 (0\%) \\
Gemini 3 Pro & 0/7 & 0/2583 (0\%) \\
GPT-5 & 0/7 & 0/2583 (0\%) \\
GPT-5.2 & 0/7 & 0/2583 (0\%) \\
\bottomrule
\end{tabular}
\end{table}

\begin{table}[ht]
\centering
\small
\caption{Model performance on credit agreements (10~documents).}
\label{tab:domain-credit}
\begin{tabular}{lcc}
\toprule
\textbf{Model} & \textbf{Valid} & \textbf{Pass Rate} \\
\midrule
GPT-5 & 10/10 & \textbf{113/130 (86.9\%)} \\
Gemini 3 Flash & 10/10 & 111/130 (85.4\%) \\
Gemini 3 Pro & 10/10 & 110/130 (84.6\%) \\
GPT-5.2 & 10/10 & 105/130 (80.8\%) \\
Claude Sonnet 4.5 & 0/10 & 0/130 (0\%) \\
Claude Opus 4.5 & 0/10 & 0/130 (0\%) \\
\bottomrule
\end{tabular}
\end{table}

\begin{table}[ht]
\centering
\small
\caption{Model performance on professional resumes (7~documents).}
\label{tab:domain-resume}
\begin{tabular}{lcc}
\toprule
\textbf{Model} & \textbf{Valid} & \textbf{Pass Rate} \\
\midrule
Claude Opus 4.5 & 5/7 & \textbf{52/217 (24.0\%)} \\
Claude Sonnet 4.5 & 5/7 & 46/217 (21.2\%) \\
Gemini 3 Pro & 5/7 & 46/217 (21.2\%) \\
Gemini 3 Flash & 5/7 & 45/217 (20.7\%) \\
GPT-5.2 & 5/7 & 43/217 (19.8\%) \\
GPT-5 & 1/7 & 8/217 (3.7\%) \\
\bottomrule
\end{tabular}
\end{table}

\begin{table}[ht]
\centering
\small
\caption{Model performance on research papers (6~documents).}
\label{tab:domain-research}
\begin{tabular}{lcc}
\toprule
\textbf{Model} & \textbf{Valid} & \textbf{Pass Rate} \\
\midrule
Claude Sonnet 4.5 & 5/6 & \textbf{47/96 (49.0\%)} \\
Gemini 3 Flash & 5/6 & 46/96 (47.9\%) \\
Claude Opus 4.5 & 2/6 & 16/96 (16.7\%) \\
GPT-5.2 & 1/6 & 6/96 (6.2\%) \\
Gemini 3 Pro & 1/6 & 5/96 (5.2\%) \\
GPT-5 & 0/6 & 0/96 (0\%) \\
\bottomrule
\end{tabular}
\end{table}

\begin{table}[ht]
\centering
\small
\caption{Model performance on sports results (5 documents).}
\label{tab:domain-sports}
\begin{tabular}{lcc}
\toprule
\textbf{Model} & \textbf{Valid} & \textbf{Pass Rate} \\
\midrule
Gemini 3 Flash & 5/5 & \textbf{11/60 (18.3\%)} \\
Claude Opus 4.5 & 5/5 & 8/60 (13.3\%) \\
Claude Sonnet 4.5 & 5/5 & 8/60 (13.3\%) \\
Gemini 3 Pro & 5/5 & 8/60 (13.3\%) \\
GPT-5.2 & 5/5 & 8/60 (13.3\%) \\
GPT-5 & 2/5 & 2/60 (3.3\%) \\
\bottomrule
\end{tabular}
\end{table}

\begin{table*}[ht]
\centering
\small
\caption{Per-document token statistics. Input tokens are Vision tokens (raw image representation). Compression ratio = Input $\div$ Output.}
\label{tab:token-stats}
\begin{tabular}{llrrrr}
\toprule
\textbf{Domain} & \textbf{Document} & \textbf{Pages} & \textbf{Input (Vision)} & \textbf{Output} & \textbf{Ratio} \\
\midrule
\multirow{5}{*}{Sports Results} 
& ma\_2023\_sw\_M-table1 & 2 & 4,634 & 468 & 9.9 \\
& ma\_2023\_sw\_M-table2 & 3 & 6,951 & 250 & 27.8 \\
& ma\_2023\_sw\_M-table3 & 3 & 6,951 & 268 & 25.9 \\
& ma\_2023\_sw\_M-table4 & 3 & 6,951 & 385 & 18.1 \\
& ma\_2023\_sw\_M-table5 & 4 & 9,268 & 551 & 16.8 \\
\midrule
\multirow{6}{*}{Research Papers}
& NIPS-1989 (backprop) & 9 & 22,810 & 815 & 28.0 \\
& FlashAttention-3 & 22 & 55,726 & 4,245 & 13.1 \\
& Dimensionality reduction survey & 35 & 81,095 & 7,551 & 10.7 \\
& RAG survey & 18 & 45,594 & 13,403 & 3.4 \\
& VLM survey & 22 & 55,726 & 20,766 & 2.7 \\
& LLM survey & 144 & 364,752 & 91,261 & 4.0 \\
\midrule
\multirow{7}{*}{SEC 10-K/10Q}
& ADP 10-Q FY25Q2 & 43 & 99,588 & 2,734 & 36.4 \\
& Nike 10-Q FY25Q2 & 48 & 111,168 & 3,045 & 36.5 \\
& THO 10-Q FY25Q2 & 53 & 122,748 & 2,779 & 44.2 \\
& McKesson 10-Q FY25Q2 & 58 & 134,328 & 2,824 & 47.6 \\
& WDC 10-Q FY25Q2 & 59 & 136,644 & 3,003 & 45.5 \\
& Cisco 10-Q FY25Q2 & 79 & 182,964 & 3,137 & 58.3 \\
& Dell 10-Q FY25Q2 & 82 & 189,912 & 2,538 & 74.8 \\
\midrule
\multirow{10}{*}{Credit Agreements}
& Adobe (2000) & 97 & 245,701 & 306 & 803 \\
& Disney (2022) & 102 & 236,232 & 510 & 463 \\
& Boeing (2003) & 110 & 278,630 & 360 & 774 \\
& 3M (2019) & 112 & 283,696 & 654 & 434 \\
& Expel (2023) & 123 & 284,868 & 365 & 780 \\
& Cisco (2007) & 126 & 319,158 & 725 & 440 \\
& BKRF (2020) & 140 & 324,240 & 551 & 588 \\
& Trimble (2022) & 143 & 331,188 & 434 & 763 \\
& IBM (2019) & 197 & 499,001 & 798 & 625 \\
& Amazon (2014) & 218 & 552,194 & 480 & 1,150 \\
\midrule
\multirow{7}{*}{Resumes}
& Resume-Finance & 1 & 2,533 & 432 & 5.9 \\
& Resume-Legal & 1 & 2,533 & 521 & 4.9 \\
& Resume-Med & 1 & 2,533 & 680 & 3.7 \\
& Resume-IT & 1 & 2,533 & 388 & 6.5 \\
& Resume-Marketing & 3 & 7,599 & 861 & 8.8 \\
& Resume-Academic01 & 7 & 17,731 & 2,996 & 5.9 \\
& Resume-Academic02 & 7 & 17,731 & 2,861 & 6.2 \\
\bottomrule
\end{tabular}
\end{table*}

\section{Evaluation Configs}
\label{app:eval-configs}

Each field in the schema specifies an \texttt{evaluation\_config} that determines how extracted values are compared against gold annotations (see Table~\ref{tab:presets}). The following shows one example of each metric type as used in the ExtractBench schemas.

\paragraph{string\_exact} Exact string match. Used for timestamps and codes.
{\footnotesize
\begin{verbatim}
"time": {"type": "string",
         "evaluation_config": "string_exact"}
\end{verbatim}
}

\paragraph{string\_case\_insensitive} Case-folded match. Used for units and enums.
{\footnotesize
\begin{verbatim}
"unit": {"type": "string",
         "evaluation_config": "string_case_insensitive"}
\end{verbatim}
}

\paragraph{string\_fuzzy} Levenshtein similarity. Used for entity names.
{\footnotesize
\begin{verbatim}
"administrative_agent": {"type": "string",
                         "evaluation_config": "string_fuzzy"}
\end{verbatim}
}

\paragraph{string\_semantic} LLM-based semantic equivalence. Used for free-text fields.
{\footnotesize
\begin{verbatim}
"title": {"type": "string",
          "evaluation_config": "string_semantic"}
\end{verbatim}
}

\paragraph{integer\_exact} Exact integer match. Used for years and counts.
{\footnotesize
\begin{verbatim}
"year": {"type": "integer", "minimum": 1000, "maximum": 2100,
         "evaluation_config": "integer_exact"}
\end{verbatim}
}

\paragraph{number\_exact} Exact numeric match. Used for monetary amounts.
{\footnotesize
\begin{verbatim}
"amount": {"type": "number",
           "evaluation_config": "number_exact"}
\end{verbatim}
}

\paragraph{number\_tolerance} Numeric match within a relative margin. Used for reported financial values.
{\footnotesize
\begin{verbatim}
"value": {"type": "number",
          "evaluation_config": {"metric_id": "number_tolerance",
                                "params": {"tolerance": 0.001}}}
\end{verbatim}
}

\paragraph{boolean\_exact} Exact boolean match. Used for binary flags.
{\footnotesize
\begin{verbatim}
"isCurrent": {"type": "boolean",
              "evaluation_config": "boolean_exact"}
\end{verbatim}
}

\paragraph{array\_llm} LLM-based semantic alignment of arrays. Used for variable-length lists.
{\footnotesize
\begin{verbatim}
"authors": {"type": "array",
            "evaluation_config": "array_llm",
            "items": { ... }}
\end{verbatim}
}

\section{Extraction Prompt}
\label{app:prompt}

The following zero-shot prompt was used for all baseline evaluations in prompt mode:

\begin{lstlisting}[basicstyle=\ttfamily\small, breaklines=true, frame=single]
Using the JSON template as a guideline, extract all
the required information from {document_name} document.

JSON Template:
{schema_str}

Please return ONLY valid JSON that conforms to this
schema. Do not include any explanatory text before
or after the JSON.
\end{lstlisting}

\end{document}